\documentclass[10pt,twocolumn,letterpaper]{article}

\usepackage{iccv}
\usepackage{times}
\usepackage{epsfig}
\usepackage{graphicx}
\usepackage{amsmath}
\usepackage{amssymb}

\usepackage[accsupp]{axessibility} % Improves PDF readability for those with disabilities.
\usepackage{microtype}
\usepackage{subfigure}
\usepackage{booktabs}
\usepackage{multirow}
\usepackage{dsfont}
\usepackage{diagbox}

\usepackage[pagebackref=true,breaklinks=true,letterpaper=true,colorlinks,bookmarks=false]{hyperref}

\iccvfinalcopy % *** Uncomment this line for the final submission

\def\iccvPaperID{10212} % *** Enter the ICCV Paper ID here

\ificcvfinal\fi

\begin{document}

%%%%%%%%% TITLE
% \title{Why Does Batch Normalization Increase Accuracy at the Cost of Adversarial Robustness?}
\title{Batch Normalization Increases Adversarial Vulnerability and Decreases Adversarial Transferability: A Non-Robust Feature Perspective}

\author{Philipp Benz\thanks{Equal contribution}\\
{\tt\small pbenz@kaist.ac.kr}
\and 
Chaoning Zhang$^*$\\
{\tt\small chaoningzhang1990@gmail.com}
\and 
In So Kweon\\
{\tt\small iskweon77@kaist.ac.kr}\\
% \and
\and 
Korea Advanced Institute of Science and Technology (KAIST)
}

\maketitle
\ificcvfinal\thispagestyle{empty}\fi

%%%%%%%%% ABSTRACT
\begin{abstract}
Batch normalization (BN) has been widely used in modern deep neural networks (DNNs) due to improved convergence. BN is observed to increase the model accuracy while at the cost of adversarial robustness. There is an increasing interest in the ML community to understand the impact of BN on DNNs, especially related to the model robustness. This work attempts to understand the impact of BN on DNNs from a non-robust feature perspective. Straightforwardly, the improved accuracy can be attributed to the better utilization of useful features. It remains unclear whether BN mainly favors learning robust features (RFs) or non-robust features (NRFs). Our work presents empirical evidence that supports that BN shifts a model towards being more dependent on NRFs. To facilitate the analysis of such a feature robustness shift, we propose a framework for disentangling robust usefulness into robustness and usefulness. Extensive analysis under the proposed framework yields valuable insight on the DNN behavior regarding robustness, \eg DNNs first mainly learn RFs and then NRFs. The insight that RFs transfer better than NRFs, further inspires simple techniques to strengthen transfer-based black-box attacks. \footnote{Code: \url{https://github.com/phibenz/adversarial_ml.research}}
\end{abstract}

\section{Introduction}
Batch normalization (BN)~\cite{ioffe2015batch} has been considered as a milestone technique in the development of deep neural networks (DNNs) pushing the frontier in computer vision due to improved convergence. Numerous works have attempted to understand the impact of BN on DNNs from various perspectives. In contrast to previous works, investigating why (or how) BN helps the optimization~\cite{santurkar2018does,awais2020revisiting}, our work focuses on the \textit{consequence} of such enhanced optimization, especially on the model robustness. Our work is not the first one to study BN and robustness together. Most of the previous works are focusing on the covariate shift~\cite{benz2020revisiting,schneider2020improving,xie2019intriguing,xie2020adversarial}. For example,~\cite{benz2020revisiting} adapts the BN statistics to improve the model robustness against common corruptions. On the contrary, our work studies BN by focusing on its impact on adversarial robustness from the non-robust feature perspective.

\begin{figure}
\centering
    \includegraphics[width=0.9\linewidth]{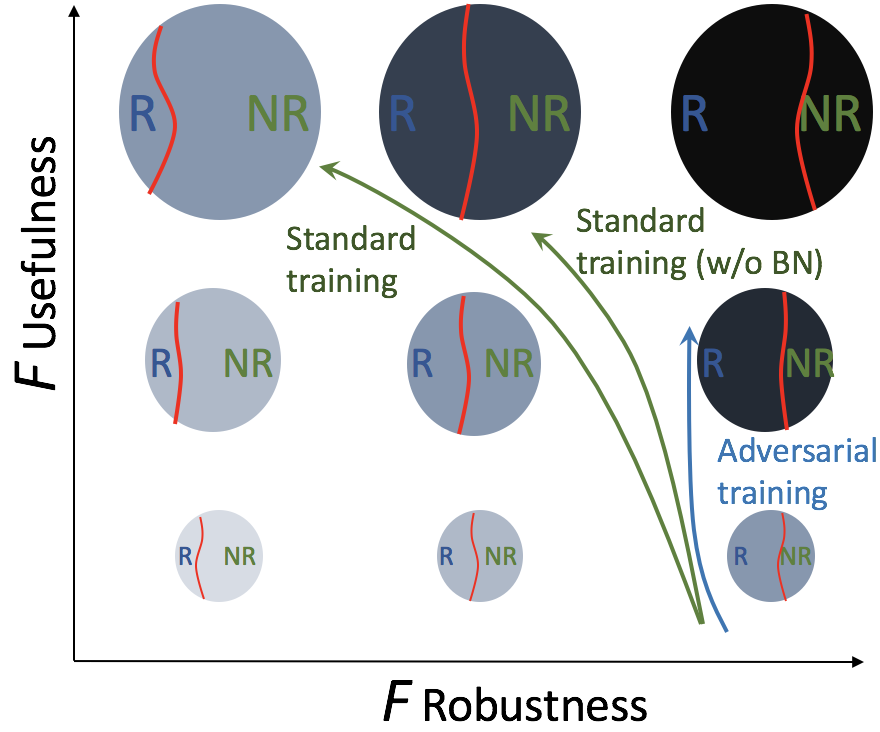}
    \caption{Schematic of disentangling $F$ usefulness and robustness with ball color representing robust usefulness, \ie the darker, the more robustly useful. Ball size indicates usefulness while the red line divides RFs and NRFs.}
    \label{fig:disentangling}
\end{figure}

We evaluate the behavior of models with and w/o BN on multiple datasets in Table~\ref{tab:influence_bn}. As expected, BN improves the clean accuracy, \ie accuracy on clean images. However, this comes at the cost of lower robust accuracy, \ie accuracy on adversarial images~\cite{szegedy2013intriguing}. Straightforwardly, the DNN can be seen as a set of useful features, consisting of robust features (RFs) and non-robust features (NRFs)~\cite{ilyas2019adversarial}, and the improved accuracy can be roughly interpreted as BN facilitating utilization of more useful features. Yet, it remains unclear whether BN mainly favors learning RFs or NRFs. Our empirical investigation shows that BN and other normalization variants all increase adversarial vulnerability in standard training, suggesting \textit{BN shifts the model to rely more on NRFs than RFs for classification.} Our claim is further corroborated by the analysis of corruption robustness and feature transferability.

With the above empirical evidence supporting our main claim that BN shifts the model towards being more dependent on NRFs, it is still necessary yet non-trivial to define and measure such a feature robustness shift. Inspired by~\cite{ilyas2019adversarial}, with a classifier DNN defined as a feature set $F$, we propose a framework, as shown in Figure~\ref{fig:disentangling}, for disentangling $F$ robust usefulness into $F$ robustness and $F$ usefulness. Following~\cite{ilyas2019adversarial}, $F$ usefulness and $F$ robust usefulness can be measured by clean accuracy and robust accuracy, respectively. $F$ usefulness can be seen as the amount of total useful features, indicated by the ball size and $F$ robustness indicates the ratio of RFs to NRFs (see Figure~\ref{fig:disentangling}). Conceptually, $F$ robustness is orthogonal to $F$ usefulness. The core difference between our feature analysis framework and that in~\cite{ilyas2019adversarial} lies in the disentangled $F$ robustness which can be utilized to measure how much BN shifts the model towards NRFs. In practice, however, it is very difficult to directly measure $F$ robustness. Inspired by~\cite{moosavi2019robustness,qin2019adversarial} demonstrating a positive correlation between robustness and local linearity, we propose a metric termed \textit{Local Input Gradient Similarity} (LIGS) (see Sec.~\ref{sec:framework}), measuring the local linearity of a DNN as an indication for $F$ robustness. Admittedly, comparing the clean accuracy and robust accuracy also sheds some light on the $F$ robustness, however, they are heavily influenced by the dimension of usefulness. Measuring LIGS provides direct evidence on how BN influences the robustness of learned $F$, which facilitates analysis under the above framework.

Such analysis yields insight on the DNN behavior regarding robustness. On a normal dataset, introducing BN (or IN/LN/GN) into the DNN consistently reduces $F$ robustness, which naturally explains their induced lower robust accuracy. We investigate and compare the behaviour of models trained on a dataset that mainly has either RFs or NRFs, which shows that NRFs are difficult to learn w/o BN, suggesting that BN is essential for learning NRFs. Further investigation on the dataset with RFs and NRFs cued for conflicting labels reveals that the model learns first RFs and then NRFs, and the previous learned RFs can be partially forgotten while the model learns NRFs in the later stage. The proposed framework is not limited for analyzing the impact of BN, and we also analyze other network structures and optimization factors. Interestingly, we find that most of them have no significant influence on $F$ robustness indicated by the LIGS metric, leaving BN (and other normalization variants) among our investigated factors as the only one that have significant influence on the shift towards more NRFs. One practical use case of our key findings is to boost transferable attacks. We demonstrate that a substitute model w/o BN outperforms its counterpart with BN and that early-stopping the training of the substitute model can also boost transferable attacks.

\begin{table}
\caption{Comparison of models with and w/o BN on accuracy and robustness.~\cite{galloway2019batch} reports a similar phenomenon.}
\label{tab:influence_bn}
    % \small
    \centering
    \setlength\tabcolsep{2.5pt}
    \scalebox{0.9}{
    \begin{tabular}{c cc cc cc cc cc}
        \toprule 
        \multirow{2}{*}{}  & \multirow{2}{*}{Network} & \multirow{2}{*}{Acc} & PGD $l_2$ & PGD $l_\infty$ & CW $l_2$ & CW $l_\infty$ \\
                 &  &  & $0.25$ & $1/255$ & $0.25$ & $1/255$ \\
        \midrule
        \parbox[t]{2mm}{\multirow{8}{*}{\rotatebox[origin=c]{90}{ImageNet}}}
        & VGG16 (None)    & $71.59$ & $15.55$ & $1.79$ & $16.66$ & $0.23$ \\
        & VGG16 (BN)      & $73.37$ & $6.04$  & $0.55$ & $6.82$  & $0.02$ \\
        & VGG19 (None)    & $72.38$ & $16.52$ & $2.18$ & $17.46$ & $0.30$ \\
        & VGG19 (BN)      & $74.24$ & $6.94$  & $0.69$ & $7.66$  & $0.03$ \\
        & ResNet18 (None) & $66.51$ & $30.44$ & $1.24$ & $30.43$ & $0.93$ \\
        & ResNet18 (BN)   & $70.50$ & $16.79$ & $0.14$ & $17.40$ & $0.07$ \\
        & ResNet50 (None) & $71.60$ & $28.00$ & $2.17$ & $28.26$ & $0.88$ \\
        & ResNet50 (BN)   & $76.54$ & $19.50$ & $0.53$ & $20.19$ & $0.19$ \\
        \midrule
        \parbox[t]{2mm}{\multirow{4}{*}{\rotatebox[origin=c]{90}{SVHN}}}
        & VGG11 (None)   & $95.42$ & $63.91$ & $83.20$ & $64.64$ & $83.24$ \\
        & VGG11 (BN)     & $96.27$ & $51.22$ & $77.50$ & $51.13$ & $77.61$ \\
        & VGG16 (None)   & $95.76$ & $62.24$ & $82.76$ & $62.97$ & $82.92$ \\
        & VGG16 (BN)     & $96.43$ & $52.90$ & $80.24$ & $52.88$ & $79.93$ \\
        \midrule
        \parbox[t]{2mm}{\multirow{6}{*}{\rotatebox[origin=c]{90}{CIFAR10}}}
        & VGG11 (None) & $90.06$ & $51.30$ & $70.47$ & $51.75$ & $70.40$ \\
        & VGG11 (BN)   & $92.48$ & $39.31$ & $63.87$ & $39.04$ & $63.85$ \\
        & VGG16 (None) & $91.89$ & $34.01$ & $63.18$ & $34.37$ & $63.46$ \\
        & VGG16 (BN)   & $93.7$ & $28.61$  & $56.05$ & $24.01$ & $54.58$ \\
        & ResNet50 (None)   & $92.15$ & $29.24$ & $49.33$ & $17.09$ & $49.24$ \\
        & ResNet50 (BN)     & $95.6$  & $9.15$  & $36.37$ & $8.72$  & $36.64$ \\
        \bottomrule
    \end{tabular}
    }
\end{table}

\begin{table*}[t]
\centering
\caption{Influence of various normalization techniques on accuracy (left$/$) and robustness ($/$right).}
\label{tab:influence_normalization}
    \scalebox{0.9}{
    \begin{tabular}{c cc cc cc cc cc}
        \toprule 
        Data  &  Network   &  None & BN & IN & LN & GN \\
        \midrule
        \multirow{2}{*}{SVHN} 
        & VGG11 & $95.42 / 63.91$ & $96.27 / 51.22$ & $95.89 / 45.82$ & $96.29 / 56.77$ & $96.30 / 56.37$ \\
        & VGG16 & $95.76 / 62.24$ & $96.43 / 52.90$ & $96.64 / 47.43$ & $96.18 / 59.55$ & $96.21 / 59.50$ \\
        \midrule
        \multirow{3}{*}{CIFAR10}
         & VGG11     & $90.06 / 51.30$ & $92.48 / 39.31$ & $88.42 / 31.38$ & $90.54 / 42.41$ & $90.68 / 39.43$ \\
         & VGG16     & $91.89 / 34.02$ & $93.70 / 28.61$ & $90.73 / 13.44$ & $92.51 / 28.92$ & $92.83 / 26.73$ \\
         & ResNet50  & $92.15 / 29.24$ & $95.60 / 9.15$  & $93.40 / 10.80$ & $90.37 / 7.24$  & $92.61 / 6.43$  \\
         \midrule
        \multirow{2}{*}{ImageNet}
        & ResNet18  & $66.51 / 30.44$ & $70.50 / 16.79$ & $63.14 / 14.29$ & $68.36 / 19.72$ & $69.02 / 19.76$ \\
        & ResNet50  & $71.60 / 28.00$ & $76.54 / 19.50$ & $67.97 / 13.65$ & $71.08 / 17.38$ & $74.69 / 20.34$ \\
        \bottomrule
    \end{tabular}
    }
\end{table*}

\section{Related Work}

\textbf{Adversarial Vulnerability and Transferability.}
Adversarial examples~\cite{szegedy2013intriguing, goodfellow2014explaining} have attracted significant attention in machine learning, which raises concern for improving the model robustness~\cite{carlini2017towards}. The cause of adversarial vulnerability has been explored from different perspectives, such as local linearity~\cite{goodfellow2014explaining}, input high-dimension~\cite{gilmer2018adversarial,shafahi2018adversarial,mahloujifar2019curse}, limited sample~\cite{schmidt2018adversarially,tanay2016boundary}, boundary tilting~\cite{tanay2016boundary}, test error in noise~\cite{fawzi2016robustness,ford2019adversarial,cohen2019certified}, etc. The cause of adversarial vulnerability has recently been attributed to highly predictive yet brittle NRFs~\cite{ilyas2019adversarial}. ~\cite{ilyas2019adversarial} proposes a feature analysis framework that discusses feature usefulness and robust usefulness that can be measured by (clean) accuracy and robust accuracy, respectively. Despite efforts to bridge their gap~\cite{zhang2019theoretically}, it is widely recognized that there is a trade-off between them~\cite{tsipras2018robustness}. Our framework proposes another dimension of feature robustness that is orthogonal to feature usefulness. On the other hand, one intriguing property of adversarial examples are their transferability, \ie\ adversarial examples generated on a substitute model is also often effective in attacking an unknown target model~\cite{kurakin2016adversarial}. Complementary to existing techniques~\cite{dong2018boosting,xie2019improving,dong2019evading}, our finding from the NRF perspective results in simple techniques that boost transferability.

\textbf{Batch normalization and beyond.} 
Since the advent of BN~\cite{ioffe2015batch}, numerous works have investigated it from various perspectives. BN performs normalization along batch dimension to reduce covariate shift, resulting in improved convergence~\cite{ioffe2015batch}. The stochasticity of the batch statistics also serves as a regularizer and improves generalization~\cite{luo2018towards}. However, the property of batch dependence limits the applicability of BN when large batch size is impractical~\cite{ba2016layer}, or there is a domain change~\cite{rebuffi2017learning}. To avoid such an issue related to the batch dimension, several alternative normalization techniques have been proposed to exploit the channel dimension, such as layer normalization (LN)~\cite{ba2016layer} in transformers and Instance normalization (IN)~\cite{vedaldi2016instance} in style transfer. LN and IN can be seen as two special cases of Group normalization (GN) in~\cite{wu2018group}. Complementary to~\cite{galloway2019batch} showing BN increases adversarial vulnerability, our work finds that LN/IN/GN mirrors the same trend. Recently, Xie \etal\ show that BN might prevent the model from obtaining strong robustness when clean examples are included in adversarial training due to the two-domain hypothesis~\cite{xie2019intriguing} and that the usage of an auxiliary batch norm for adversarial examples can improve image recognition~\cite{xie2020adversarial}. A similar approach has been adopted in~\cite{jiang2020robust} for adversarial contrastive learning. Recently~\cite{benz2020revisiting,schneider2020improving} show that covariate shift adaptation at the inference stage can enhance the robustness against common corruptions. 

\section{RFs \vs.\ NRFs: Which Side does BN Favor?}
\subsection{Background and Motivation}
\label{background_motivation}

\textbf{Reason vs Effect.} BN~\cite{ioffe2015batch} is widely adopted due to its improved convergence and the community has attempted to understand how BN helps the optimization. BN was first motivated to reduce the internal covariate shift (ICS)~\cite{ioffe2015batch}, while~\cite{santurkar2018does} claims that reducing ICS does \textit{not} help optimization, instead, the improved optimization is mainly attributed to BN smoothing the optimization landscape. One recent work~\cite{awais2020revisiting} revisits ICS and refutes the claim in~\cite{santurkar2018does} and suggests that reducing the ICS is actually the reason. Overall, the mechanism of why BN improves the optimization remains unclear and probably no clear consensus will be found in the near future. In contrast to previous works~\cite{santurkar2018does,awais2020revisiting} investigating the \textit{reason} of the improved optimization, our work focuses on the \textit{effect}, more specifically, on the adversarial robustness.
Given that a DNN learns a set of features~\cite{ilyas2019adversarial}, the improved optimization is expected to lead to a model with more useful features, consequently improving the accuracy, as expected. Its side effect of increasing adversarial vulnerability is worth an investigation. Inspired by~\cite{ilyas2019adversarial}, our work focuses on the NRF perspective. 

\textbf{RFs vs.\ NRFs.}
\emph{Feature} is one key concept in computer vision and the past few years have witnessed a shift from hand-crafted features~\cite{lowe2004distinctive,dalal2005histograms} to DNNs intrinsically extracting features~\cite{krizhevsky2012imagenet,he2016deep}. Despite different interpretations of how DNN works, there is a belief that a classification DNN can be perceived as a function utilizing useful features~\cite{ilyas2019adversarial}. Specifically,~\cite{ilyas2019adversarial} defines a \textit{feature} to be a function mapping from the input space $\mathcal{X}$ to real numbers, \ie $f: \mathcal{X}\rightarrow \rm I\!R$. A feature $f$ is $\rho$-useful ($\rho>0$) if it is correlated with the true label in expectation, \ie ${\rm I\!E_{(x,y) \sim \mathcal{D}}} [y \cdot f(x)] \ge \rho$. Given a $\rho$-useful feature $f$, robust features (RFs) and non-robust features (NRFs) are formally defined as follows:
\begin{itemize}
    \item \textit{RF:} a feature $f$ is robust if there exists a $\gamma>0$ for it to be $\gamma$-robustly useful under some specified set of valid perturbations $\Delta$, \ie ${\rm I\!E_{(x,y) \sim \mathcal{D}}} [\inf\limits_{\delta \in \Delta(x)} y \cdot f(x+\delta)] \ge \gamma$.
    \item \textit{NRF:} a feature $f$ is non-robust if $\gamma>0$ does not exist.
\end{itemize}

\textbf{Conjecture.} With the above definition~\cite{ilyas2019adversarial} to dichotomize features, we conjecture that BN shifts the model to rely more on NRFs instead of RFs.

\subsection{Empirical evidence}
\label{sec:evidence}
The observation in Table~\ref{tab:influence_bn} constitutes evidence for our conjecture, which can be corroborated as follows. First, BN simply normalizes the DNN intermediate feature layers; thus if our conjecture is correct, other normalization techniques (such as LN, IN, and GN) are also likely to mirror the same behavior. Second, with the link between adversarial robustness and corruption robustness~\cite{ford2019adversarial}, our 
claim can be more convincing if the corruption robustness analysis also supports it.

\textbf{Adversarial robustness.}
Table~\ref{tab:influence_normalization} shows that the phenomenon of increased adversarial vulnerability is not limited to BN but also occurs for IN, LN, and GN. Overall, except for the result of ResNet50 on CIFAR10, IN consistently achieves the lowest robust accuracy. We suspect that this can be attributed to the prior finding~\cite{vedaldi2016instance,nam2018batch} that IN excludes style information by performing instance-wise normalization. The style information is likely RFs (note that changing style, \eg color, normally requires large pixel intensity change), thus IN discarding style can result in the least robust model. 

\begin{figure}
    \centering
    \includegraphics[width=\linewidth]{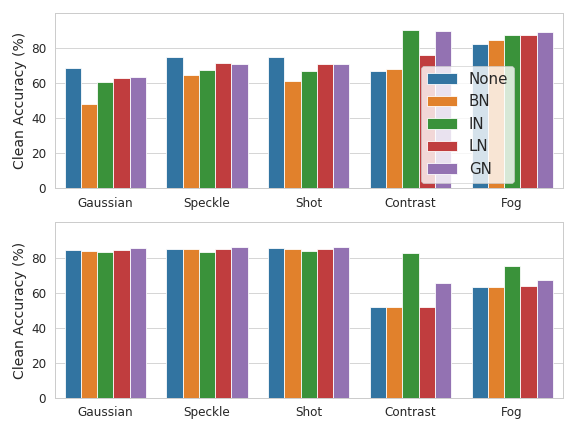}
    \caption{Corruption robustness of VGG16 with \emph{standard training} (top) and \emph{adversarial training} (bottom).}
    \label{fig:corruption_img}
\end{figure}

Ford~\etal~\cite{ford2019adversarial} revealed that adversarial training (and Gaussian data augmentation) significantly improve the robustness against noise corruptions, \ie Gaussian/Speckle/Shot, while decreasing the robustness against contrast and fog, which is confirmed in Figure~\ref{fig:corruption_img}. Following~\cite{ilyas2019adversarial}, a standard model is perceived to learn a sufficient amount of NRFs, while a robust model (robustified through adversarial training) mainly has RFs. Perceiving from the feature perspective, the following explanation arises: noise corruptions mainly corrupt the NRFs while contrast and fog mainly corrupt the RFs. Our explanation from the feature perspective echoes with prior explanation from a frequency perspective~\cite{yin2019fourier}. We discuss their link in the supplementary.

\textbf{Corruption robustness.}
The model without normalization are more robust to noise corruptions than their counterparts with normalization while a reverse trend is observed for fog and contrast. Given our explanation, this contrasting behavior suggests that the models with normalization learn more NRFs instead of RFs. Another observation from Figure~\ref{fig:corruption_img} is that IN leads to extra high-robustness against contrast corruption, suggesting less IN is the least dependent on RFs in this context. This aligns well with the previous result that the model with IN is generally the least robust.

\section{Framework for disentangling usefulness and robustness}
\label{sec:framework}
Following~\cite{ilyas2019adversarial}, we define a DNN classifier as a set of features, \ie $F = \{f\}$. The definitions of $f$ usefulness and robust usefulness in Sec.~\ref{background_motivation} can be readily extended to $F$. 
\begin{itemize}
    \item $F$ usefulness: $F$ is $\rho$-useful ($\rho>0$) if it is correlated with the true label in expectation, \ie ${\rm I\!E_{(x,y) \sim \mathcal{D}}} [y \cdot F(x)] \ge \rho$; 
    \item $F$ robust usefulness: $F$ is $\gamma$-robustly useful if there exists a $\gamma>0$ under some specified set of valid perturbations $\Delta$, \ie ${\rm I\!E_{(x,y) \sim \mathcal{D}}} [\inf\limits_{\delta \in \Delta(x)} y \cdot F(x+\delta)] \ge \gamma$.
\end{itemize}
For being orthogonal to usefulness, we can not trivially define $F$ robustness by measuring its correlation with the true label in expectation. 
With a locally quadratic approximation, prior work~\cite{moosavi2019robustness} provided theoretical evidence of a strong relationship between robustness and local linearity. Thus, with $\nabla l(x,y)$ denoting the partial gradient of the CE loss $l$ with respect to the $x$ input, we define $F$ robustness as follows. 
\begin{itemize}
    \item $F$ robustness: A feature set $F$ is $\beta$-robust if the local linearity is larger than $\beta$ ($\beta>0$), \ie ${\rm I\!E_{(x,y)\sim D, \nu \sim \Delta }} \left[ sim (\nabla l(x,y), \nabla l(x+\nu,y)) \right] \ge \beta$. 
\end{itemize}
The local linearity indicated by the similarity ($sim$) between $\nabla (l(x,y)$ and $\nabla l(x+\nu,y)$ can be represented in different forms, such as calculating the norm of their difference~\cite{moosavi2019robustness}. We adopt the cosine similarity~\cite{zhang2020understanding} to quantify this similarity as:
\begin{equation}
\mathbf{E}_{(x,y)\sim D, \nu \sim \Delta} \left[ \frac{ \nabla l(x,y) \cdot \nabla l(x+\nu,y)}{\Vert \nabla l(x,y) \Vert \cdot \Vert \nabla l(x+\nu,y) \Vert} \right].
\label{eq:local_linearity}
\end{equation}
The adopted metric indicates similarity (or linearity) between the original and locally perturbed input gradient and is thus termed \textit{Local Input Gradient Similarity} (LIGS). For additional justification for adopting this metric, refer to the supplementary. Nonetheless, metrics other than LIGS might also be appropriate. 

\textbf{Perturbation choice of LIGS.} We investigate the influence of perturbation type by setting $\nu$ to Gaussian noise, uniform noise, FGSM perturbation, and PGD perturbation. Among all the chosen types of perturbation, we observe a general trend that the LIGS decreases with training, and consistently the LIGS w/o BN is higher than that with BN. Unless specified, we sample the $\nu$ from a Gaussian distribution to measure the LIGS in this work. Additional details and results can be found in the supplementary.

\textbf{Relation to prior works.}
The primary motivation of adopting LIGS in this work is to define and quantify the $F$ robustness. Directly maximizing the local linearity as a new regularizer has been shown to improve adversarial robustness on par with adversarial training~\cite{moosavi2019robustness}. A similar finding has also been shown in~\cite{qin2019adversarial}. Note that ``adversarial robustness" mostly refers to ``robust usefulness" instead of solely ``robustness". To avoid confusion, we highlight that \emph{$F$ robustness} is orthogonal to usefulness. Contrary to prior works~\cite{moosavi2019robustness, qin2019adversarial}, which improve ``adversarial robustness" by investigating (and establishing) the link between robust usefulness with local linearity, we adopt the local linearity as a measure of ``robustness". By definition, local linearity does not imply usefulness because it is not related to the correlation with the true label. Nonetheless, their observation that maximizing local linearity can help improve robust usefulness (measured by robust accuracy), can be seen as a natural consequence of increasing $F$ robustness. 

\textbf{Interpretation and relationship.} 
Informally but intuitively, the usefulness of $F$ can be perceived as the number of features if we assume that each feature is equally useful for classification; and the robustness of $F$ can be seen as the ratio of RFs to NRFs in $F$. This is illustrated schematically in Figure~\ref{fig:disentangling}, where a DNN located in the top right region has high robust usefulness, \ie high robust accuracy, indicating the model learns sufficient features and among them, a high percentage belongs to RFs. A low robust accuracy can be caused by either low $F$ usefulness or low $F$ robustness. Figure~\ref{fig:disentangling} also shows the difference between standard training (green) and adversarial training (blue). Both start from the state of high $F$ robustness and low $F$ usefulness; compared with standard training, adversarial training eventually leads to a model of higher $F$ robustness and lower $F$ usefulness. For standard training, removing BN also increases $F$ robustness. By definition, $F$ robustness, $F$ usefulness, and $F$ robust usefulness can be measured by LIGS, clean accuracy, and robust accuracy, respectively. The schematic illustration in Figure~\ref{fig:disentangling} aligns well with the results in Figure~\ref{fig:acc_cossim_normalization}. 

\section{Disentangling usefulness and robustness of model features}
\label{sec:disentangling}
With the above evidence to corroborate that BN shifts the model to rely more on NRFs, it would be desirable to have a metric to measure ``pure" robustness independent of usefulness. Given both RFs/NRFs are useful and their core difference lies in robustness, such a metric is crucial for providing direct evidence on the shift towards NRFs by showing a lower ``pure" robustness. Moreover, the LIGS trend during the training stage also sheds light on the learned order of features, \ie from RFs to NRFs or vice versa. Evaluating adversarial robustness by robust accuracy demonstrates how robustly useful the model features are. Thus, disentangling robust usefulness into usefulness and robustness provides a better understanding of adversarial robustness. 

\begin{figure}[!htbp]
\centering
    \includegraphics[width=\linewidth]{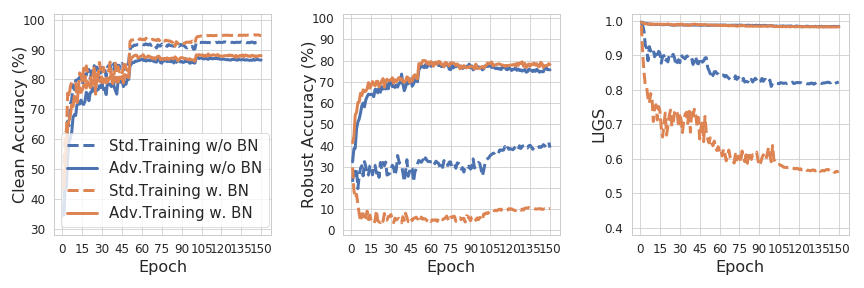}
    \caption{Trend of clean accuracy, robust accuracy, LIGS with ResNet18 on CIFAR10. 
    }
    \label{fig:acc_cossim_normalization}
\end{figure}

The overall trend in Figure~\ref{fig:acc_cossim_normalization} shows that robust accuracy is influenced by both clean accuracy and LIGS. For example, for adversarial (adv.) training, the LIGS stays close to 1 during the entire training stage, and the robust accuracy is highly influenced by the clean accuracy. For standard (std.) training, however, the LIGS is much lower, leading to a much smaller robust accuracy despite slightly higher clean accuracy. The influence of BN is mainly observed on LIGS. During the entire training stage, BN leads to a significantly lower LIGS, consequently lower robust accuracy. 

As (standard) training evolves, the LIGS value decreases, \ie the feature robustness decreases, suggesting the model relies more on NRFs as training evolves. The influence of BN in adv.\ training, however, is limited. Here, only BN on CIFAR10 is reported. We provide more results with IN/LN/GN and results on ImageNet in the supplementary. The results mirror the trend in Figure~\ref{fig:acc_cossim_normalization}.

\textbf{On the role of BN in adversarial training.} With standard training, we find that BN increases adversarial vulnerability. To improve robustness, adversarial training is one of the most widely used methods. The authors of~\cite{xie2019intriguing} showed that BN might prevent networks from obtaining strong robustness in adversarial training. However, this is only true when clean images are utilized in the training and the reason is attributed to the two-domain hypothesis. For standard adversarial training~\cite{madry2017towards} with only adversarial images, as shown in Figure~\ref{fig:acc_cossim_normalization}, BN is found to have no influence on LIGS as well as robust accuracy. This is reasonable because adversarial training explicitly discards NRFs. 

\begin{figure}[!htbp]
    \centering
        \includegraphics[width=\linewidth]{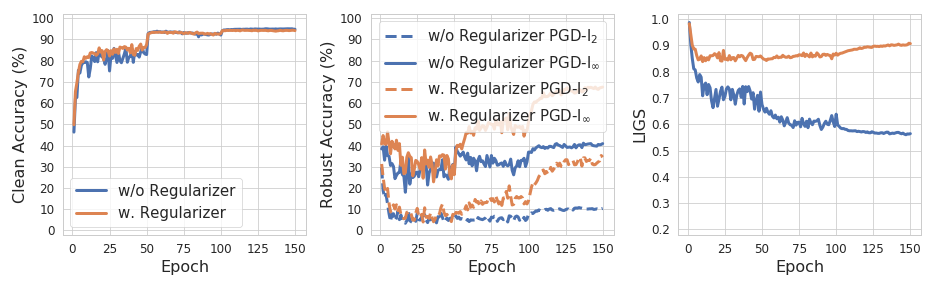}
        \caption{Effect of regularizing LIGS.}
        \label{fig:regularization}
\end{figure}

\textbf{Regularization of LIGS}. The above results show that the robust accuracy and LIGS are linked. To verify the link between them, we use LIGS as a regularizer during training. The results in Figure~\ref{fig:regularization} confirm that increasing LIGS through regularization improves the robust accuracy by a large margin despite a small influence on clean accuracy. 

\subsection{Training on a dataset of disentangled RFs and NRFs}
Note, that by default the experiment setup is the same by only changing the variable of interest (\eg testing with and without BN). Training on a dataset of disentangled RFs and NRFs with BN as the control variable highlights the effect of BN on them while excluding mutual influence. 

\begin{figure}[!htbp]
    \centering
        \includegraphics[width=\linewidth]{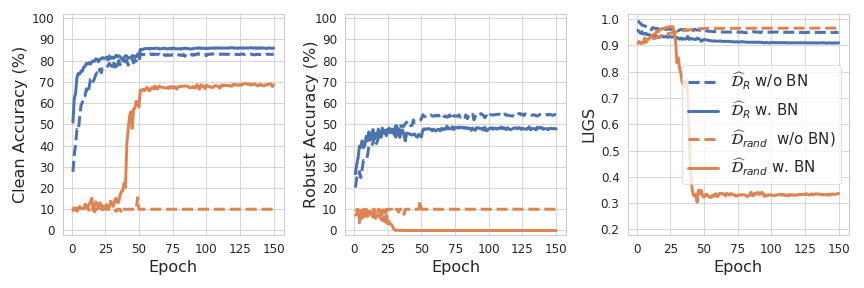}
        \caption{Analysis of BN with ResNet18 on datasets of disentangled features.}
        \label{fig:compare_resnet18_none_bn}
\end{figure}

\textbf{Disentangling RFs and NRFs.}
Following the procedure of~\cite{ilyas2019adversarial} we extract $\widehat{\mathcal{D}}_{R}$, $\widehat{\mathcal{D}}_{NR}$ and $\widehat{\mathcal{D}}_{rand}$ (Description in the supplementary). Note that  $\widehat{\mathcal{D}}_{R}$ mainly (if not exclusively) has RFs, while $\widehat{\mathcal{D}}_{rand}$ only has NRFs. $\widehat{\mathcal{D}}_{NR}$ has both RFs and NRFs (see the supplementary for results on $\widehat{\mathcal{D}}_{NR}$). 
Here, to demonstrate the effect of BN on either NRFs or NRFs, we report the results trained on $\widehat{\mathcal{D}}_{R}$ and $\widehat{\mathcal{D}}_{rand}$ in Figure~\ref{fig:compare_resnet18_none_bn}, where the clean accuracy and robust accuracy results echo the findings in~\cite{ilyas2019adversarial}. There are two major observations regarding the LIGS result. First, the LIGS on $\widehat{\mathcal{D}}_{R}$ is very high (more than 0.9), which explains why a model (normally) trained on $\widehat{\mathcal{D}}_{R}$ has relatively high robust accuracy, while the LIGS on $\widehat{\mathcal{D}}_{rand}$ eventually becomes very low because $\widehat{\mathcal{D}}_{rand}$ only has NRFs. 
Second, w/o BN, the model is found to not converge on $\widehat{\mathcal{D}}_{rand}$, leading to 10\% accuracy, \ie equivalent to random guess. The model with BN starts to converge (achieving an accuracy of higher than 10\%) after around 25 epochs and the LIGS is observed to increase before the model starts converging. This suggests that the model is learning features that are robust yet hardly useful. This ``warmup" phenomenon is not accidental and repeatedly happens with different random training seeds. After the model starts to converge, the LIGS quickly plummets to a low value.

\textbf{Training on a dataset of conflicting RFs and NRFs.} 
In the original dataset, $\mathcal{D}$, abundant RFs and NRFs co-exist and the model learns both for classification. It is interesting to understand the order of the learned features, \ie from RFs to NRFs or vice versa, as well their influence on each other. The decreasing trend of LIGS in Figure~\ref{fig:acc_cossim_normalization} suggests that the model learns mainly RFs first. Here, we provide another evidence with the metric of clean accuracy. In the $\mathcal{D}$, RFs and NRFs are cued for the same classification, thus no insight can be deduced from the clean accuracy. To this end, we design a dataset $\widehat{\mathcal{D}}_{Conflict}$ of conflicting RFs and NRFs. Specifically, we exploit the generated $\widehat{\mathcal{D}}_{R}$ of target class $t+1$ as the starting images and generate the NRFs of the target class $t$. In other words, in the $\widehat{\mathcal{D}}_{Conflict}$ RFs are cued for class $t+1$ while NRFs are cued for class $t$. 

\begin{figure}[!htbp]
    \centering
        \includegraphics[width=0.9\linewidth]{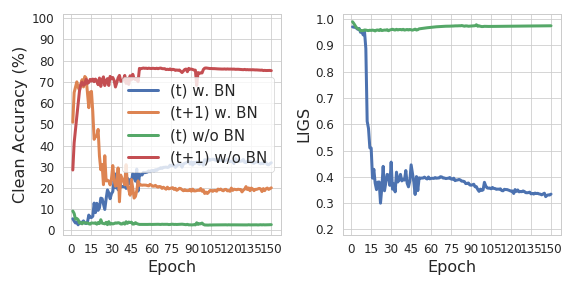}
        \caption{Analysis on dataset with Conflicting RFs and NRFs.}
        \label{fig:conflict}
\end{figure}

Figure~\ref{fig:conflict} shows that with BN the clean accuracy aligned with RFs increases significantly in the first few epochs and peaks around $80\%$ followed by a sharp decrease, while the accuracy aligned with NRFs slowly increases until saturation. It supports that the model learns from RFs to NRFs. Eventually, the accuracy aligned with NRFs surpasses that aligned with RFs, indicating the model forgets most of the first learned RFs during the later stage. W/o BN, we find that the model in the whole stage learns RFs while ignoring NRFs. It clearly shows that BN is crucial for learning NRFs, which naturally explains why BN shifts the models towards learning more NRFs. We also discuss the results of $\widehat{\mathcal{D}}_{det}$~\cite{ilyas2019adversarial} in the supplementary. 

\subsection{Exploration beyond (batch) normalization}
\begin{figure*}[!htbp]
    \centering
        \includegraphics[width=0.16\linewidth]{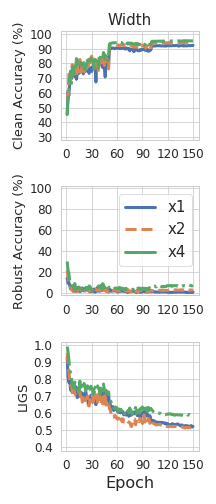}
        \includegraphics[width=0.16\linewidth]{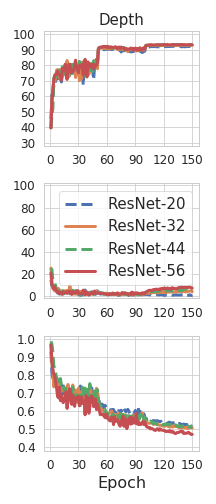}
        \includegraphics[width=0.16\linewidth]{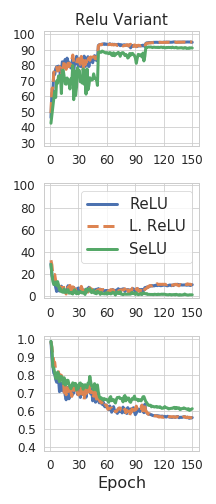}
        \includegraphics[width=0.16\linewidth]{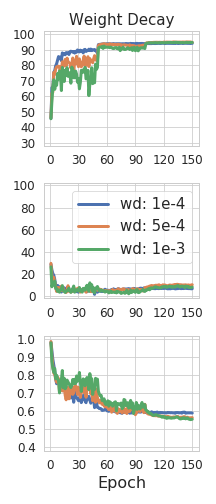}
        \includegraphics[width=0.16\linewidth]{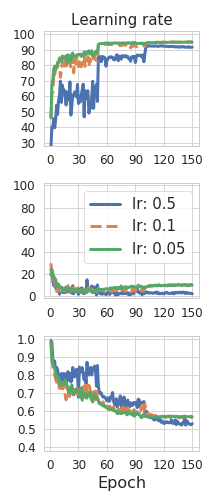}
        \includegraphics[width=0.16\linewidth]{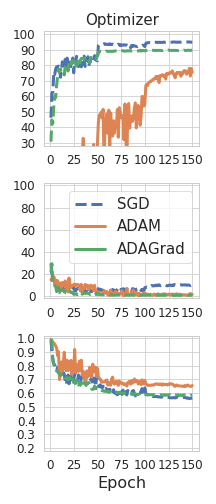}
        \caption{Influence of other factors on the behavior or DNN.}
        \label{fig:other_influences}
\end{figure*}

\textbf{Network structure factors and optimization factors.} Besides normalization, other factors could influence the DNN behavior, especially concerning its robustness. We study two categories of factors: (a) structure category including network width, network depth, and ReLu variants; (b) optimization category including weight decay, initial learning rate, and optimizer. The results are presented in Figure~\ref{fig:other_influences}. 
We find that most studied factors have no significant influence on the LIGS. Increasing network width and depth can increase or decrease the LIGS, respectively, but by a small margin. No visible difference between ReLU and Leaky ReLU can be observed, while SeLU leads to a slightly higher LIGS with lower clean accuracy. Some candidates from the optimization category are found to influence $F$ robustness differently in the early and later stages of training. High weight decay leads to higher LIGS in the early stages of training and slightly lower in the end. A higher initial learning rate, such as $0.5$, results in higher LIGS in the early training stage, but eventually leads to lower LIGS. For both weight decay and initial learning rate, an opposite trend of lower/higher in the early/later stage is observed with clean accuracy. SGD optimizer and ADAGrad show similar behavior on LIGS, ADAM leads to slightly higher LIGS. Their influence on clean accuracy is more significant.

\section{On the link between why BN favors NRFs and how BN helps optimization.}
As discussed in Sec.~\ref{background_motivation}, there are two major conflicting views on how BN helps optimization: (a) smoothing the optimization landscape~\cite{santurkar2018does} \vs\ (b) reducing ICS~\cite{awais2020revisiting}. View (a) and view (b) hold exactly the opposite claims against each other. On the other hand, why BN shifts towards NRFs also remains unclear. The shift towards NRFs is caused by the improved optimization of BN (note that both views agree that BN improves the optimization). Thus, (1) why BN shifts towards NRFs and (2) how BN improves the optimization are essentially the same, at least highly correlated, problems. We suggest future works investigate problem (1) and problem (2) jointly. Here, we perform a trial attempt in this direction.

\begin{figure}[!htbp]
    \centering
        \includegraphics[width=0.8\linewidth]{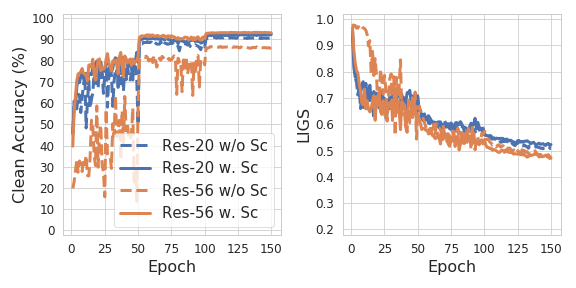}
        \includegraphics[width=0.8\linewidth]{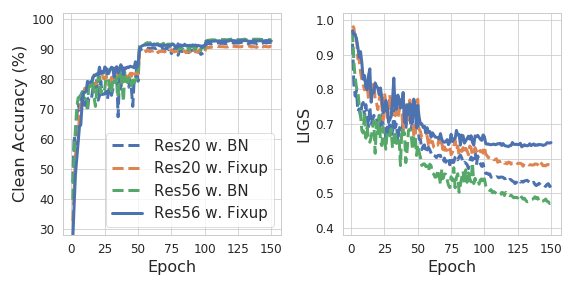}
        \caption{Effect of shortcut (top) and FixupIni (bottom) on model.}
        \label{fig:comparison_shortcut}
\end{figure}

Given our observation that the model w/o BN cannot converge on a dataset with only NRFs and the wide belief that BN stabilizes training, we are wondering about a potential link between training stability and $F$ robustness. ResNet shortcut also stabilizes/accelerates training~\cite{he2016deep}, thus we investigate whether it reduces $F$ robustness. Figure~\ref{fig:comparison_shortcut} shows that shortcut has trivial influence on LIGS with ResNet20. For a much deeper ResNet56, removing the shortcut has a significant influence on LIGS in the early stage of training, however, eventually, the influence also becomes marginal. Fixup initialization (FixupIni) is introduced in~\cite{zhang2019fixup} to replace the BN in ResNets. We compare their influence on the model and observe that their difference in clean accuracy is trivial, while BN leads to lower LIGS than FixupIni. Overall, it shows increasing training stability does not necessarily lead to lower $F$ robustness. If stabilizing training does not necessarily result in a shift towards NRFs, it is likely that view (a) does not hold because it is deduced based on the observation that BN leads to a more predictive and stable gradient. Moreover, IN/LN/GN is also found to improve the optimization as well as a shift towards NRFs. If view (a) holds, likely, IN/LN/GN will also lead to a more predictive and stable gradient. Following~\cite{santurkar2018does}, we visualize the gradient predictiveness and find that they do not lead to strong gradient stability as BN (See the supplementary). One common thing between BN/IN/LN/GN is that they all reduce ICS and the evidence we find supports view (b). Note that the authors of this work do not have any interest in conflict with the above two views and just objectively present the evidence we collect. The authors also have no intention to claim that view (b) is the final reason and welcome future works to present with more supportive or contradicting evidence. 

\section{Implications of our findings for improving adversarial transferability}
\label{sec:implications}
Recent works~\cite{salman2020adversarially,utrera2020adversarially,terzi2020adversarial} show that robust models are more suitable for downstream tasks for transfer learning, suggesting models that contain more robust features transfer better across tasks. One natural conjecture is that such models might also be more suitable for being used as the substitute model for generating transferable adversarial examples across models. This conjecture is confirmed by our preliminary finding that adversarial examples generated on the adversarially trained models transfer better to normal models. However, typical adversarial training requires more computation resources~\cite{madry2017towards,zhang2019theoretically}. Our findings regarding the RFs/NRFs can be utilized in normal training for boosting the adversarial transferability. 

\begin{table}[t]
\centering
\caption{Influence of BN on the transferability. Results on ImageNet with various baselines: I-FGSM~\cite{kurakin2016adversarial}, MI-FGSM~\cite{dong2018boosting}, DI-FGSM~\cite{xie2019improving} and TI-FGSM~\cite{dong2019evading}.}
    \setlength\tabcolsep{2.5pt}
    \scalebox{0.82}{
    \begin{tabular}{ccc| cc cc cc cc| c}
        \toprule 
        & Source & BN  & RN50 & DN121 & VGG19 & RN152 & MN-V2 & I-V3 & Avg \\ \midrule
        \multirow{4}{*}{I}
        & VGG19    & Y & 47.3 & 49.5 & 100 & 32.3 & 58.8 & 20.1 & 51.3 \\
        & VGG19    & N & 65.4 & 65.7 & 98.0 & 48.1 & 77.6 & 32.1 & 64.5 \\
        & RN50     & Y & 100  & 80.1 & 71.6 & 86.2 & 73.4 & 34.2 & 74.2 \\
        & RN50     & N & 98.6 & 94.3 & 87.0 & 95.5 & 94.4 & 72.1 & 90.3 \\
        \midrule
        \multirow{4}{*}{MI}
        & VGG19    & Y & 60.7 & 65.3 & 100 & 44.3 & 70.1 & 36.7 & 62.9 \\
        & VGG19    & N & 73.8 & 76.4 & 98.5 & 58.8 & 83.7 & 47.5 & 73.1 \\
        & RN50     & Y & 100 & 88.8 & 81.9 & 92.8 & 83.0 & 50.7 & 82.9 \\
        & RN50     & N & 98.9 & 95.4 & 88.7 & 95.5 & 96.2 & 78.5 & 92.2 \\
        \midrule
        \multirow{4}{*}{DI}
        & VGG19    & Y & 65.4 & 68.0 & 100 & 46.3 & 75.2 & 28.9 & 64.0 \\
        & VGG19    & N & 77.1 & 74.6 & 99.0 & 56.8 & 85.5 & 37.4 & 71.7 \\
        & RN50     & Y & 100 & 98.1 & 96.9 & 97.9 & 94.4 & 59.8 & 91.2 \\
        & RN50     & N & 99.4 & 99.1 & 95.8 & 98.1 & 98.8 & 90.3 & 96.9 \\
        \midrule
        \multirow{4}{*}{TI}
        & VGG19    & Y & 57.9 & 58.2 & 100.0 & 43.5 & 70.5 & 30.5 & 60.1 \\
        & VGG19    & N & 71.3 & 70.9 & 97.7 & 53.7 & 79.0 & 40.9 & 68.9 \\
        & RN50     & Y & 100 & 82.4 & 75.4 & 88.6 & 77.1 & 40.3 & 77.3 \\
        & RN50     & N & 98.7 & 95.0 & 87.0 & 95.7 & 95.2 & 77.6 & 91.5 \\
        \bottomrule
    \end{tabular}
    }
\label{tab:transferability_imagenet}
\end{table}

\begin{table}[t]
\centering
\caption{Influence of BN on the transferability.  Results on CIFAR10 with various baselines: I-FGSM~\cite{kurakin2016adversarial}, MI-FGSM~\cite{dong2018boosting}. Full results with DI-FGSM~\cite{xie2019improving} and TI-FGSM~\cite{dong2019evading} are in the supplementary.}
    \setlength\tabcolsep{2.5pt}
    \scalebox{0.82}{
    \begin{tabular}{c cc|cccccc|c}
        \toprule 
        & Source & BN & AlexN & VGG16 & RN50 & DN & RNext & WRN & Avg \\ \midrule
        \multirow{4}{*}{I}
        & VGG16    & Y & 28.7 & 100$^*$ & 85.7 & 81.7 & 84.7 & 83.3 & 77.4 \\
        & VGG 16   & N & 39.5 & 99.8 & 99.6 & 98.0 & 98.8 & 98.9 & 89.1 \\
        & ResNet18 & Y & 24.7 & 73.3 & 80.7 & 80.0 & 83.8 & 85.7 & 71.4  \\
        & ResNet18 & N & 41.5 & 99.6 & 99.7 & 98.3 & 99.4 & 98.9 & 89.6  \\
        \midrule 
        \multirow{4}{*}{MI}
        & VGG16    & Y & 34.4 & 100$^*$ & 93.9 & 91.1 & 92.2 & 92.7 & 84.1 \\
        & VGG 16   & N & 44.1 & 99.7 & 99.2 & 97.2 & 98.0 & 98.4 & 89.4 \\
        & ResNet18 & Y & 28.7 & 86.9 & 90.1 & 88.3 & 90.8 & 92.7 & 79.6 \\
        & ResNet18 & N & 45.1 & 99.1 & 99.0 & 96.6 & 98.2 & 97.9 & 89.3 \\
        \bottomrule
    \end{tabular}
    }
\label{tab:transferability_cifar10}
\end{table}

One takeaway from this work is that BN shifts the model to utilize more NRFs than RFs. A normal model (with BN by default) is used in the existing approaches~\cite{dong2018boosting,xie2019improving,dong2019evading}. Given RFs transfer better, we experiment with a substitute model w/o BN on ImageNet (see Table~\ref{tab:transferability_imagenet}) and CIFAR10 (see Table~\ref{tab:transferability_cifar10}). We observe that on a wide range of DNN architectures, the substitute models w/o BN transfer significantly better than their counterparts. The results here also provide additional evidence that BN indeed shifts the model to rely more on NRFs. Recently, Normalization-Free networks~\cite{brock2021high} have been introduced. We leave an investigation of their robustness and transferability for future investigations.

\begin{figure}[!htbp]
    \centering
        \includegraphics[width=0.8\linewidth]{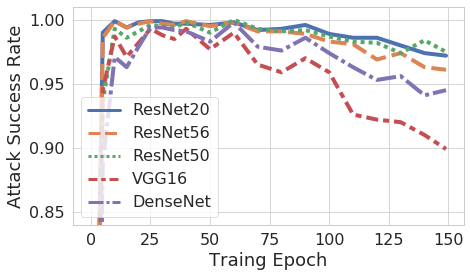}
        \caption{Performance of a substitute ResNet18 measured across different training epochs on 5 black-box models.}
        \label{fig:transferability_epochs}
\end{figure}

Another interesting finding of this is that the DNN mainly first learns RFs and then learns NRFs. To get a substitute model with more RFs, a straightforward idea inspired by this finding is to train the substitute model with an early stop. By default, existing methods train a substitute model trained with full epochs. We report the transferability performance for substitute models at different epochs in Figure~\ref{fig:transferability_epochs}. We observe that the transferability performance increases very sharply in the early epochs, and decreases gradually in the later epochs. The results demonstrate that early stopping indeed helps significantly improve transferability. 

Transfer-based black-box attack is a vibrant and competitive research field~\cite{dong2018boosting,xie2019improving,dong2019evading,gao2020patch}, and our proposed two techniques are expected to be complementary to most of the existing techniques. More transferability results are shown in the supplementary and we highlight that our findings have important implications for understanding adversarial transferability from the NRF perspective as well as provide direct insight with simple yet effective techniques for boosting transferable black-box attacks.

\vspace{-2mm}
\section{Conclusion}
BN and other normalization variants increase adversarial vulnerability. We attribute the reason to a shift of the model to rely more on NRFs, for which we provide empirical evidence from both adversarial robustness and corruption robustness analysis. We propose a framework for disentangling the usefulness and robustness of model features. With the disentangled interpretation, we find that the model learns first RFs and then NRFs because RFs are essential for training the model in the early stage, and that BN is crucial for learning NRFs. The reason why BN shifts the model towards NRFs and how BN helps the optimization are essentially the same problem. A joint analysis of the observed phenomena shows that the current evidence supports the view of reducing ICS between two conflicting views. Our findings also provide a new understanding of adversarial transferability from the NRF perspective as well as inspire two simple yet effective ideas for boosting transferable attacks.

\section*{Acknowledgements}
This work was supported in part by the Institute of Information and Communications Technology Planning and Evaluation (IITP) Grant funded by the Korea Government (MSIT) (Artificial Intelligence Innovation Hub) under Grant 2021-0-02068.

{\small
\bibliographystyle{ieee_fullname}
\bibliography{bib_mixed}
}

\appendix
\clearpage

\section{Relation to frequency perspective}
\label{sec:link_frequency}
Our work focuses on the feature perspective~\cite{ilyas2019adversarial} to analyze the model robustness. A Fourier perspective on robustness is introduced in~\cite{yin2019fourier}. With the analysis of corruption analysis in Sec.~\ref{sec:evidence}, our explanation from the feature perspective is ``noise corruptions mainly corrupt the NRFs while contrast and fog mainly corrupt the RFs". Their explanation from the frequency perspective can be summarized as: noise corruptions mainly corrupt the high-frequency (HF) features while contrast and fog mainly corrupt the low-frequency (LF). These two explanations align well with each other in the sense that NRFs are widely recognized to have HF property, which motivated the exploration of several defense methods~\cite{das2018shield,liu2019feature}. Moreover, our work is the first to demonstrate that the model learns the order from RFs to NRFs. Meanwhile, it has been shown in~\cite{xu2019frequency} that the model learns first LF component then HF component, which aligns well with our finding by perceiving NRFs having HF properties.

\section{Experimental setup}
\subsection{Setup for training models in Sec.~\ref{sec:evidence}}
The models for CIFAR10 and SVHN used in Sec.~\ref{sec:evidence} were trained with SGD with the training parameters listed in Table~\ref{tab:param_normal_train}. The ResNet50 models in Table~\ref{tab:influence_bn} and Table~\ref{tab:influence_normalization} of the main manuscript were trained for $350$ epochs with an initial learning rate of $0.1$, which was decreased by a factor of $10$ at epochs $150$ and $250$, while the other parameters are the same as before. 
For ImageNet, the VGG models were obtained from the \texttt{torchvision} library, while the ResNet models are trained with the same parameters as in~\cite{he2016deep}.

\begin{table}[!htbp]
\centering
\caption{Parameters to train a standard model on CIFAR10/SVHN.}
\label{tab:param_normal_train}
    \small
    \begin{tabular}{c c}
    \toprule
    Parameter & Value  \\
    \midrule
    Learning rate              & $0.01$   \\
    Batch size                 & $128$    \\
    Weight Decay               & $0.0005$ \\
    Epochs                     & $300$    \\ 
    Learning rate decay epochs & $200$    \\
    Learning rate decay factor & $0.1$    \\
    \bottomrule
    \end{tabular}
\end{table}

\begin{table}[!htbp]
\centering
\caption{Training parameters for adversarial training for CIFAR10/SVHN.}
\label{tab:param_adv_train}
    \small
    \begin{tabular}{c c}
    \toprule
    Parameter & Value  \\
    \midrule
    Learning rate                           & $0.01$   \\
    Batch size                              & $128$    \\
    Weight Decay                            & $0.0005$ \\
    Epochs                                  & $150$    \\ 
    Learning rate decay epochs              & $100$    \\
    Learning rate decay factor              & $0.1$    \\
    PGD-variant                             & $l_2$    \\
    PGD step size ($\alpha$)                & $0.1$    \\
    PGD perturbation magnitude ($\epsilon$) & $0.5$    \\
    PGD iterations                          & $7$      \\
    \bottomrule
    \end{tabular}
\end{table}
\subsection{PGD attack for evaluating the adversarial robustness}
In Table~\ref{tab:influence_bn}, we evaluate the robustness of models with the $l_2$ and $l_\infty$ variants of the PGD-attack~\cite{madry2017towards} and  Carlini \& Wagner (CW) attack~\cite{carlini2017towards}. For the $l_2$ and $l_\infty$ attack we use $\epsilon=0.25$ and $\epsilon=1/255$ for images within a pixel range of $[0,1]$, respectively. The attacks are run for $20$ iteration steps and we calculate the step size with $2.5 \epsilon/\text{steps}$. For the CW-attack~\cite{carlini2017towards}, we follow~\cite{zhang2019defense} to adopt the PGD approach with the CW loss and the same hyper parameters as above.

The robust accuracy values in all figures in Sec.~\ref{sec:disentangling} are obtained with $l_2$-PGD as above, but for $10$ iteration steps on $1000$ evaluation samples ($100$ samples per class) to reduce computation cost.

\begin{figure}
    \centering
        \includegraphics[width=1.0\linewidth]{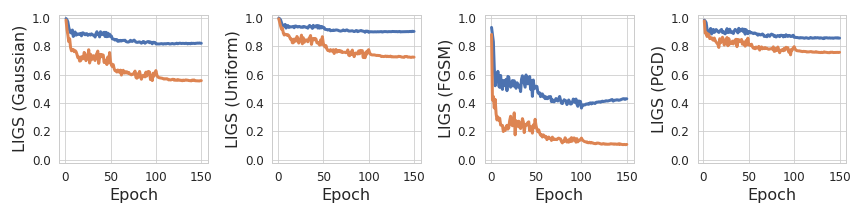}
        \caption{Trend of LIGS with different perturbations: Gaussian, Uniform, FGSM, PGD (left to right).}
        \label{fig:LIGS_noise_comparison}
\end{figure}

\subsection{LIGS metric}
\label{app:ligs}
By default, the LIGS values are calculated with $\nu$ being set to Gaussian noise with mean $\mu=0$ and standard deviation $\sigma=0.01$. In Figure~\ref{fig:LIGS_noise_comparison} $\nu$ is set to either Gaussian noise, uniform noise in the range of $[-0.01, 0,01]$, FGSM with $\epsilon=0.01$ or $l_\infty$-PGD with $\epsilon=0.01$, $7$ step iterations and a step size of $2.5\epsilon/\text{steps}$.

\begin{table}
\centering
\caption{Cross-evaluation of the features extracted from source models on target models with the baseline VGG16 for standard models.}
\label{tab:feature_transerability_standard}
    \scalebox{0.9}{
    \begin{tabular}{c cc cc cc cc cc}
        \toprule 
              \diagbox{Target}{Source} & None & BN & IN & LN & GN \\
        \midrule 
        None      & $-$    & $45.6$ & $29.6$ & $52.1$ & $45.9$ \\
        BN & \textbf{75.3} & $-$    & $35.8$ & $58.9$ & $54.3$ \\
        IN & \textbf{66.1} & $50.4$ & $-$    & $53.0$ & $61.9$ \\
        LN & \textbf{79.4} & $59.4$ & $37.9$ & $-$    & $63.4$ \\
        GN & \textbf{73.3} & $54.4$ & $43.1$ & $61.3$ & $-$ \\
        \bottomrule
    \end{tabular} 
    }
\end{table}

\begin{table}
\centering
\caption{Cross-evaluation of the features extracted from source models on target models with the baseline VGG16 for adversarially trained models.}
\label{tab:feature_transerability_adversarial}
    \scalebox{0.9}{
    \begin{tabular}{c cc cc cc cc cc}
        \toprule 
                \diagbox{Target}{Source}  & None & BN & IN & LN & GN  \\
        \midrule 
        None      & $-$ & $85.0$ & $59.8$ & $75.8$ & $65.9$ \\
        BN & $81.3$ & $-$ & $58.1$ & $73.7$ & $62.5$ \\
        IN & $75.8$ & $78.8$ & $-$ & $69.4$ & $62.9$ \\
        LN & $82.9$ & $84.5$ & $60.0$ & $-$ & $64.8$ \\
        GN & $80.7$ & $83.7$ & $63.8$ & $73.7$ & $-$ \\
        \bottomrule
    \end{tabular}
    }
\end{table}
\section{Feature Transferability} 
In Sec.~\ref{background_motivation} we formulated the conjecture, that BN shifts the model to rely more on NRFs instead of RFs and provided empirical evidence for this conjecture. Additional to the empirical evidence given in Sec.~\ref{sec:evidence}, we provide one additional piece of evidence via a feature transferability analysis. We extract the features out of the trained models as a new dataset and perform cross-evaluation on the remaining models (details in the supplementary). The results are shown in Table~\ref{tab:feature_transerability_standard}. As a control study, we perform the same analysis on adversarially trained robust models, see Table~\ref{tab:feature_transerability_adversarial}. It has been shown in~\cite{salman2020adversarially} that robust models are superior to normally trained models for transfer learning, which demonstrates that RFs can transfer better. Here, we find that features extracted from robust models (right) can transfer better than the features extracted from standard models (left). For the normally trained models, we observe that the features extracted from the model w/o normalization transfer better (indicated in bold) than those models with BN/IN/LN/GN, especially IN. Recognizing the extracted features have both RF and NRFs, our observation suggests that the models with normalization rely more on NRFs than that w/o BN.

\subsection{Extracting features as a dataset from a model}
\label{app:extracting_features}
To demonstrate feature transferability in Table~\ref{tab:feature_transerability_standard} and Table~\ref{tab:feature_transerability_adversarial}, we extract features from standard and adversarially trained models as a dataset. For robust models in Table~\ref{tab:feature_transerability_adversarial} we follow the adversarial training strategy from~\cite{madry2017towards} with $l_2$-PGD and we list the parameters in Table~\ref{tab:param_adv_train}. 
We follow the procedure and hyperparameter choices in~\cite{ilyas2019adversarial} and generate dataset $\hat{\mathcal{D}}$, given a model $C$: 
\begin{equation}
    \mathds{E}_{(x,y) \sim \hat{\mathcal{D}}} [y \cdot f(x)] =
    \begin{cases}
        \mathds{E}_{(x,y)\sim \mathcal{D}}[y \cdot f(x)] &\text{ if $f \in F_C$}\\
        0 & \text{ otherwise},
    \end{cases}
\end{equation}
where $F_C$ is the set of features utilized by $C$. The set of activations in the penultimate layer $g(x)$ corresponds to $F_C$ in the case of DNNs. Thus, to extract the robust features from $\mathcal{C}$ we perform the following optimization:
\begin{equation}
    \min_\delta ||g(x) - g(x^\prime+\delta)||_2.
\end{equation}
The optimization is performed for each sample $x$ from $\mathcal{D}$. Likewise, $x^\prime$ is drawn from $\mathcal{D}$ but with a label other than that of $x$. The optimization process is realized using the $l_2$ variant of PGD. We set the step size to $0.1$ and the number of iterations to $1000$ and we do not apply a restriction on the perturbation magnitude $\epsilon$.

\section{Influence of other normalization techniques on LIGS}
\label{sec:othernorm}
\begin{figure}[!htbp]
\centering 
    \includegraphics[width=0.32\linewidth]{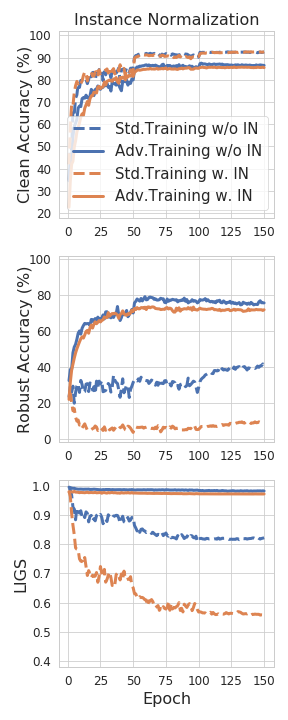}
    \includegraphics[width=0.32\linewidth]{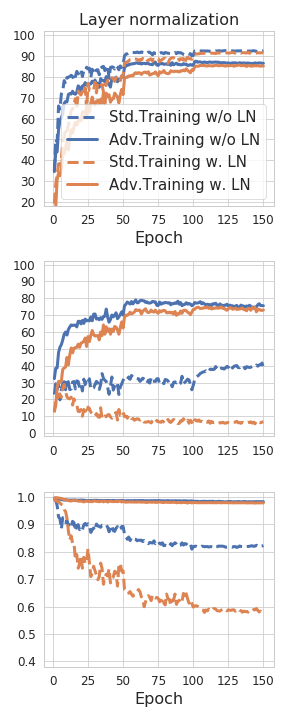}
    \includegraphics[width=0.32\linewidth]{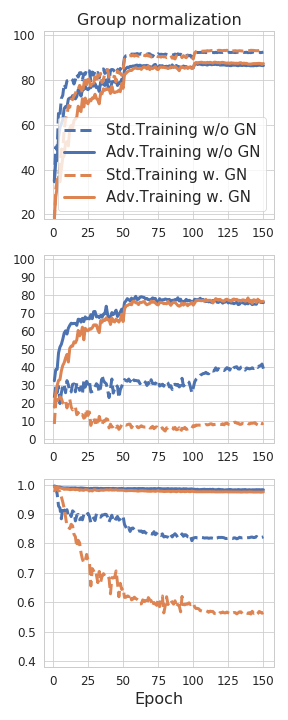}
    \caption{Trend of clean accuracy (top), robust accuracy (middle), LIGS (bottom) for ResNet18 on CIFAR10 with different normalization techniques (IN, LN, GN) applied.
    }
    \label{fig:trend_in_ln_gn_cifar}
\end{figure}

In Fig.~\ref{fig:disentangling}, we show the influence of BN on the robust accuracy and LIGS over the model training process. Additionally, Fig.~\ref{fig:trend_in_ln_gn_cifar} shows the results of repeating this experiment with IN, LN, and GN. Similar trends to those of BN are observed.

\section{Results on ImageNet with LIGS trend}
\label{sec:imagenet}
\begin{figure}[!htbp]
\centering 
    \includegraphics[width=1.0\linewidth]{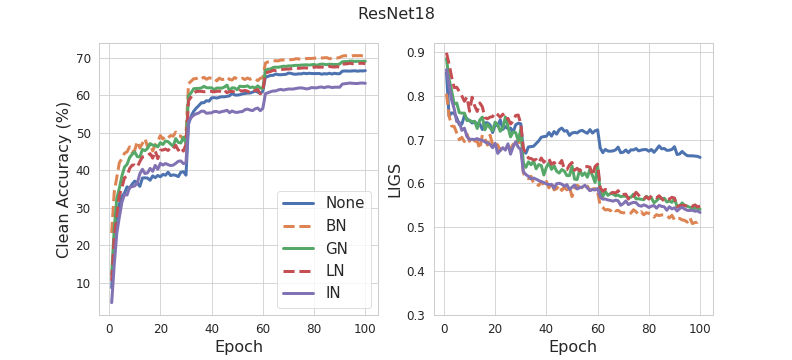}
    \includegraphics[width=1.0\linewidth]{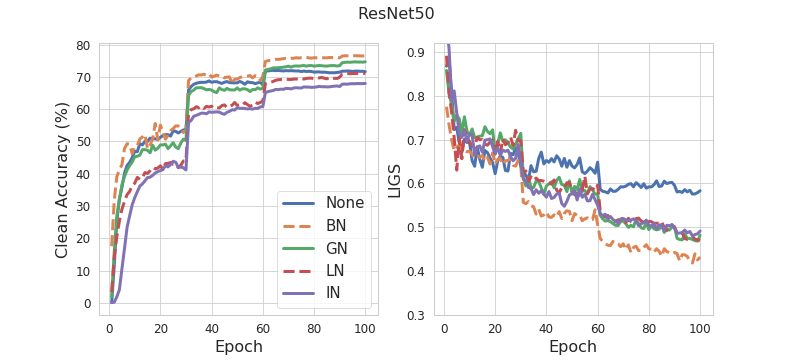}
    \caption{Comparison of different normalization techniques for ResNet18 (top) and ResNet50 (bottom) trained on ImageNet. 
    }
    \label{fig:trend_in_ln_gn_cifar_imagenet}
\end{figure}

Fig.~\ref{fig:trend_in_ln_gn_cifar_imagenet} shows the influence of normalization for models trained on ImageNet.
It can be observed that the model with IN always exhibits the lowest accuracy, while the model with BN has the highest accuracy. Similar to the results on CIFAR10, BN/IN/LN/GN consistently leads to lower LIGS. 

\section{Description of $\widehat{\mathcal{D}}_{R}$ /  $\widehat{\mathcal{D}}_{NR}$ / $\widehat{\mathcal{D}}_{rand}$ / $\widehat{\mathcal{D}}_{det}$}
\label{app:description_of_datasets}
\cite{ilyas2019adversarial} introduced a methodology to extract feature datasets from models. In particular the datasets $\widehat{\mathcal{D}}_{R}$, $\widehat{\mathcal{D}}_{NR}$, $\widehat{\mathcal{D}}_{rand}$ and $\widehat{\mathcal{D}}_{det}$ were introduced, which we will describe here briefly. 
$\widehat{\mathcal{D}}_{R}$ indicates a dataset containing mainly RFs relevant to a robust model, and $\widehat{\mathcal{D}}_{NR}$ indicates that with standard model. During the extraction of $\widehat{\mathcal{D}}_{NR}$, the magnitude $\epsilon$ was not constraint, thus $\widehat{\mathcal{D}}_{NR}$ has both RFs and NRFs. $\widehat{\mathcal{D}}_{\text{rand}}$ and $\widehat{\mathcal{D}}_{\text{det}}$ are datasets consisting of ``useful" NRFs represented through adversarial examples for a standard model. 
The target classes of $\widehat{\mathcal{D}}_{\text{rand}}$ were chosen randomly, while the ones for $\widehat{\mathcal{D}}_{\text{det}}$ were selected with an offset of $t+1$ to the original sample ground-truth class. Note that these datasets are labeled with the target class for which the adversarial example was generated. 
We follow the procedure described in~\cite{ilyas2019adversarial} and extract the datasets from a ResNet50 model. The hyperparameters used to train a model on one of the above datasets are listed in Table~\ref{tab:training_extracted_data_param}. We use SGD as an optimizer and train the models for $150$ epochs with a learning rate decrease by a factor of $10$ at epochs $50$ and $100$.

\begin{table}[t]
\centering
\caption{Hyperparameters for training the extracted datasets.}
\setlength\tabcolsep{1.5pt}
    \scalebox{0.7}{
\label{tab:training_extracted_data_param}
\begin{tabular}{l | cc cc cc cc cc}
        \toprule 
        Dataset & LR & Batch size & LR Drop & Data Aug. & Momentum & Weight Decay  \\
        \midrule 
        $\widehat{\mathcal{D}}_{R}$        & $0.01$ & $128$ & Yes & Yes & $0.9$ & $5 \cdot 10^{-4}$ \\
        $\widehat{\mathcal{D}}_{NR}$       & $0.01$ & $128$ & Yes & Yes & $0.9$ & $5 \cdot 10^{-4}$ \\
        $\widehat{\mathcal{D}}_{rand}$     & $0.01$ & $128$ & Yes & Yes & $0.9$ & $5 \cdot 10^{-4}$ \\
        $\widehat{\mathcal{D}}_{det}$      & $0.1$  & $128$ & Yes & No  & $0.9$ & $5 \cdot 10^{-4}$ \\
        $\widehat{\mathcal{D}}_{conflict}$ & $0.1$  & $128$ & Yes & No  & $0.9$ & $5 \cdot 10^{-4}$ \\
        \bottomrule
    \end{tabular}
    }
\end{table}

\begin{figure}[!htpb]
    \centering
    \includegraphics[width=1.0\linewidth]{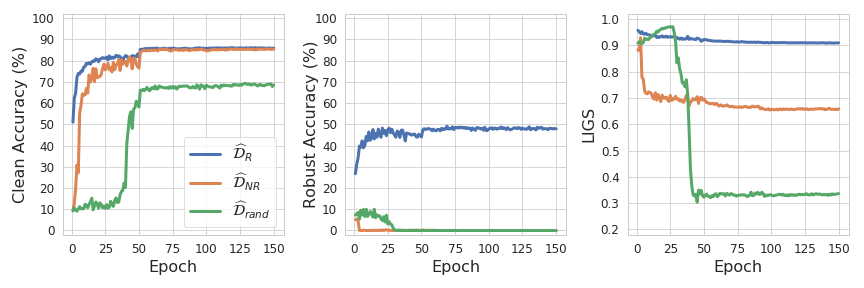}
    \caption{Comparison of $\widehat{\mathcal{D}}_{R}$, $\widehat{\mathcal{D}}_{NR}$ and $\widehat{\mathcal{D}}_{rand}$.}
    \label{fig:extracted_data_eval}
\end{figure}

Fig.~\ref{fig:extracted_data_eval} shows the trends for training ResNet18 on $\widehat{\mathcal{D}}_{R}$, $\widehat{\mathcal{D}}_{NR}$ and $\widehat{\mathcal{D}}_{rand}$. As seen before, the model trained on $\widehat{\mathcal{D}}_{R}$ achieves a relatively high LIGS, while the model trained on $\widehat{\mathcal{D}}_{rand}$ exhibits relatively low LIGS values. The LIGS values for the model trained on $\widehat{\mathcal{D}}_{NR}$ are in the middle of $\widehat{\mathcal{D}}_{R}$ and $\widehat{\mathcal{D}}_{rand}$, which is expected because $\widehat{\mathcal{D}}_{NR}$ has RFs and NRFs.

\begin{figure}[!htbp]
    \centering
    \includegraphics[width=0.8\linewidth]{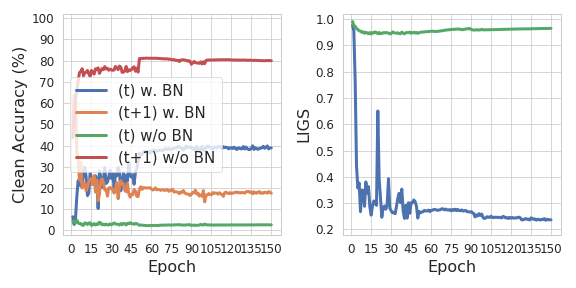}
    \caption{Result on $\widehat{\mathcal{D}}_{\text{det}}$.}
    \label{fig:ddet_standard_bg}
\end{figure}

Fig.~\ref{fig:conflict} shows the trends for training a ResNet18 on $\widehat{\mathcal{D}}_{Conflict}$, consisting of conflicting features. $\widehat{\mathcal{D}}_{Conflict}$ differs from $\widehat{\mathcal{D}}_{\text{det}}$ in that it draws $x^\prime$ from a robust dataset $\widehat{\mathcal{D}}_{R}$ instead of $\mathcal{D}$. The same experiment with $\widehat{\mathcal{D}}_{\text{det}}$ is shown in Fig.~\ref{fig:ddet_standard_bg}. The results resemble those of $\widehat{\mathcal{D}}_{Conflict}$. However, we used $\widehat{\mathcal{D}}_{Conflict}$ to avoid the influence of the NRFs in $\mathcal{D}$.

\section{Evaluating adversarial robustness with FGSM attack}
\label{sec:fgsm}
Motivated by the (local) linearity assumption,~\cite{goodfellow2014explaining} proposed the one-step FGSM attack. FGSM efficiently attacks the model but is not as effective as PGD attack~\cite{madry2017towards} because the DNN is not fully linear. With iterative nature to overcome this linearity assumption, PGD is a very strong attack and de facto benchmark standard for evaluating the adversarial robustness, due to which PGD is adopted in our work. Here, we discuss the effect of BN with FGSM. BN reduces the LIGS value, which indicates the model with BN has low local linearity. Since the success of FGSM is highly dependent on the linearity assumption, the FGSM attack is conjectured to be less effective on the model with BN than w/o BN. This conjecture is supported by the results in the supplementary.
As shown in Table~\ref{tab:fgsm}, we find that with FGSM attack, the model with BN has higher adversarial robustness than that w/o BN.

\begin{table}[!htpb]
\centering
\caption{Robust accuracy comparison of models with and w/o BN under FGSM attack.}
\scalebox{0.9}{
\label{tab:fgsm}
    \begin{tabular}{c cc cc cc cc cc}
        \toprule 
        \multirow{2}{*}{}  & \multirow{2}{*}{Network} & \multirow{2}{*}{Acc} & FGSM  \\
                 &  & & $4/255$ & \\
        \midrule
        & VGG11 (None)      & $90.06$ & $32.51$ \\
        & VGG11 (BN)        & $92.48$ & $40.86$ \\
        & VGG16 (None)      & $91.89$ & $23.26$ \\
        & VGG16 (BN)        & $93.7$  & $51.28$ \\
        & ResNet50 (None)   & $92.15$ & $28.23$ \\
        & ResNet50 (BN)     & $95.6$  & $38.07$ \\
        \bottomrule
    \end{tabular}
    }
\end{table}

\section{Additional Transferability Results}
In Section~\ref{sec:implications} we demonstrated that more strong transferable adversarial examples can be generated for models without BN. In Table~\ref{tab:transferability_imagenet_adv_trained_models} we demonstrate that adversarial examples generated on adversarially trained models transfer better than normal models. Compared to the ResNet50 model with BN, both adversarially trained models transfer better for all I-FGSM variants. Except for DI-FGSM, the adversarially trained models do also outperform the RN50 models without BN. The results for DI-FGSM and TI-FGSM for CIFAR10 are shown in Table~\ref{tab:transferability_cifar10_supp}. The results resemble the ones originally presented in Table~\ref{tab:transferability_cifar10} of the main manuscript. 

\begin{table}[!htbp]
\centering
\caption{Influence of BN on the transferability of robustly trained ResNet50 models. Results on ImageNet with various baselines: I-FGSM~\cite{kurakin2016adversarial}, MI-FGSM~\cite{dong2018boosting}, DI-FGSM~\cite{xie2019improving} and TI-FGSM~\cite{dong2019evading}.}
    \setlength\tabcolsep{2.pt}
    \scalebox{0.8}{
    \begin{tabular}{ccc| cc cc cc| c}
        \toprule 
        & Variant & BN  & RN50 & DN121 & VGG19 & RN152 & MN-V2 & I-V3 & Avg \\ \midrule
        \multirow{4}{*}{I}
        & Standard     & Y & 100  & 80.1 & 71.6 & 86.2 & 73.4 & 34.2 & 74.2 \\
        & Standard     & N & 98.6 & 94.3 & 87.0 & 95.5 & 94.4 & 72.1 & 90.3 \\
        & $L_2 = 3.0$    & Y & 98.9 & 98.6 & 94.6 & 98.3 & 98.1 & 96.5 & 97.5 \\
        & $L_\infty = 4$ & Y & 97.3 & 95.9 & 92.5 & 96.1 & 96.6 & 92.7 & 95.2 \\
        \midrule
        \multirow{4}{*}{MI}
        & Standard     & Y & 100 & 88.8 & 81.9 & 92.8 & 83.0 & 50.7 & 82.9 \\
        & Standard     & N & 98.9 & 95.4 & 88.7 & 95.5 & 96.2 & 78.5 & 92.2 \\
        & $L_2 = 3.0$    & Y & 97.6 & 97.1 & 92.9 & 96.7 & 97.6 & 94.8 & 96.1 \\
        & $L_\infty = 4$ & Y & 95.5 & 94.6 & 89.9 & 93.7 & 95.1 & 89.3 & 93.0 \\
        \midrule
        \multirow{4}{*}{DI}
        & Standard     & Y & 100 & 98.1 & 96.9 & 97.9 & 94.4 & 59.8 & 91.2 \\
        & Standard     & N & 99.4 & 99.1 & 95.8 & 98.1 & 98.8 & 90.3 & 96.9 \\
        & $L_2 = 3.0$    & Y & 98.0 & 97.5 & 91.9 & 96.3 & 97.2 & 94.3 & 95.9 \\
        & $L_\infty = 4$ & Y & 93.9 & 93.9 & 85.8 & 91.3 & 94.1 & 88.6 & 91.3 \\
        \midrule
        \multirow{4}{*}{TI}
        & Standard     & Y & 100 & 82.4 & 75.4 & 88.6 & 77.1 & 40.3 & 77.3 \\
        & Standard     & N & 98.7 & 95.0 & 87.0 & 95.7 & 95.2 & 77.6 & 91.5 \\
        & $L_2 = 3.0$    & Y & 98.5 & 98.3 & 96.2 & 97.9 & 98.6 & 95.8 & 97.5 \\
        & $L_\infty = 4$ & Y & 96.4 & 96.4 & 92.8 & 95.5 & 97.1 & 93.9 & 95.4 \\
        \bottomrule
    \end{tabular}
    }
\label{tab:transferability_imagenet_adv_trained_models}
\end{table}

\begin{table}[!htbp]
\centering
\caption{Influence of BN on the transferability.  Results on CIFAR10 with two baselines: DI-FGSM~\cite{xie2019improving}, TI-FGSM~\cite{dong2019evading}.}
    \setlength\tabcolsep{2.5pt}
    \scalebox{0.82}{
    \begin{tabular}{c cc|cccccc|c}
        \toprule 
        & Source & BN & AlexN & VGG16 & RN50 & DN & RNext & WRN & Avg \\ 
        \midrule 
        \multirow{4}{*}{DI}
        & VGG16    & Y & 28.7 & 100$^*$ & 91.8 & 89.7 & 90.9 & 90.8 & 82.0 \\
        & VGG 16   & N & 42.0 & 99.9 & 99.8 & 99.1 & 99.3 & 99.6 & 90.0 \\
        & ResNet18 & Y & 25.3 & 80.7 & 88.1 & 87.2 & 91.6 & 92.1 & 77.5 \\
        & ResNet18 & N & 41.4 & 99.6 & 99.7 & 98.5 & 99.3 & 99.1 & 89.6 \\
        \midrule 
        \multirow{4}{*}{TI}
        & VGG16    & Y & 33.9 & 100.0 & 80.8 & 74.9 & 80.7 & 78.8 & 74.9 \\
        & VGG 16   & N & 51.2 & 99.6 & 99.3 & 97.2 & 98.3 & 98.4 & 90.7 \\
        & ResNet18 & Y & 26.7 & 67.7 & 76.5 & 73.9 & 80.5 & 81.7 & 67.8 \\
        & ResNet18 & N & 53.1 & 99.1 & 99.4 & 97.0 & 98.7 & 98.0 & 90.9 \\
        \bottomrule
    \end{tabular}
    }
\label{tab:transferability_cifar10_supp}
\end{table}

\begin{figure}[!htbp]
    \centering
    \includegraphics[width=0.8\linewidth]{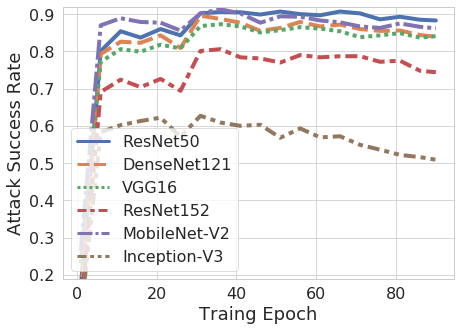}
    \caption{Performance of a substitute ResNet18 (ImageNet) model measured across different training epochs on 5 black-box models.}
    \label{fig:trans_imagenet_epochs}
\end{figure}

In Figure~\ref{fig:transferability_epochs} of the main manuscript, we demonstrated that the transferability of adversarial examples generated for a ResNet18 on CIFAR10 decreases with ongoing model training. In Figure~\ref{fig:trans_imagenet_epochs} we provide the complementary result for a ResNet18 trained on ImageNet. For ResNet18 trained on ImageNet, we observe a similar trend as on CIFAR10. The transferability initially increases and then decreases gradually during training.

\section{Visualization of the optimization landscape}
\label{app:landscape}
\begin{figure}[t]
    \centering
    \includegraphics[width=0.9\linewidth]{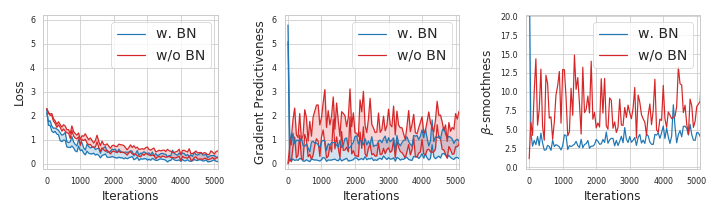}
    \includegraphics[width=0.9\linewidth]{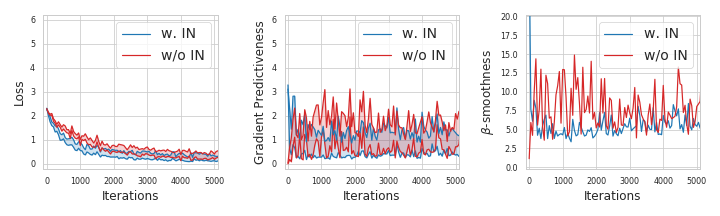}
    \includegraphics[width=0.9\linewidth]{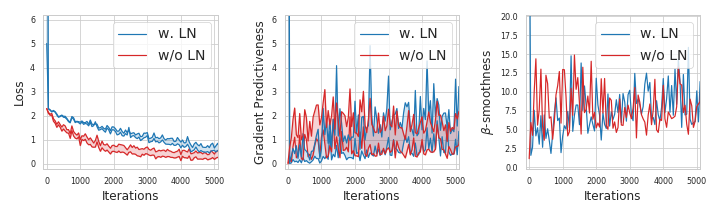}
    \includegraphics[width=0.9\linewidth]{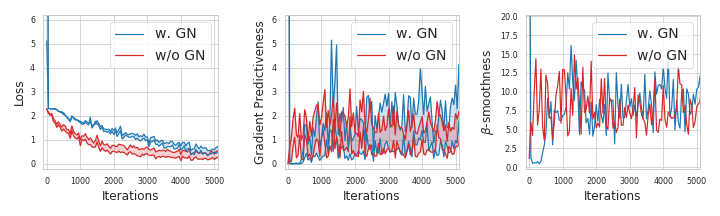}
    \caption{Optimization landscape of ResNet18 with and without normalization. 
    Variation in loss (left); $l_2$ gradient change (center); $\beta$-smoothness (right).
    }
    \label{fig:landscape_resnet18}
\end{figure}

\begin{figure}[t]
    \centering
    \includegraphics[width=0.95\linewidth]{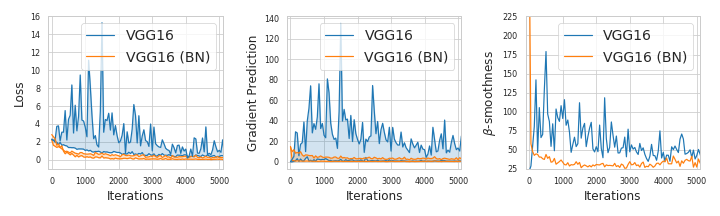}
    \includegraphics[width=0.95\linewidth]{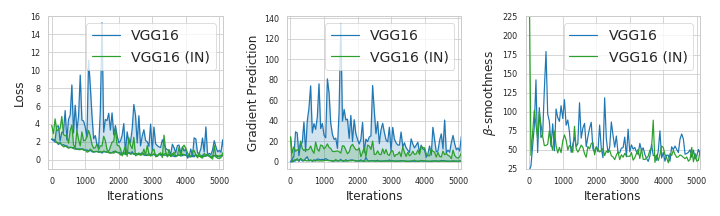}
    \includegraphics[width=0.95\linewidth]{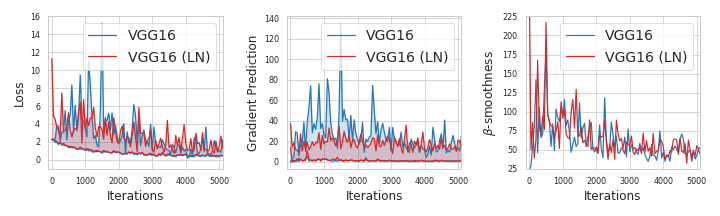}
    \includegraphics[width=0.95\linewidth]{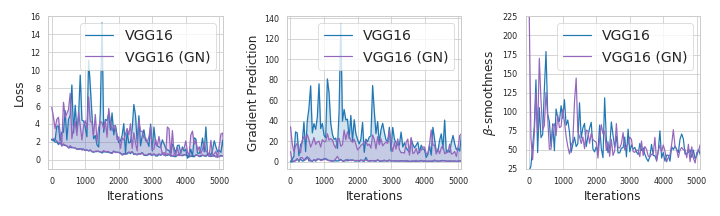}
    \caption{Optimization landscape of VGG16 with and without BN. 
    Variation in loss (left); $l_2$ gradient change (center); $\beta$-smoothness (right).
    }
    \label{fig:landscape_vgg16}
\end{figure}

\begin{figure}[t]
    \centering
    \includegraphics[width=1.0\linewidth]{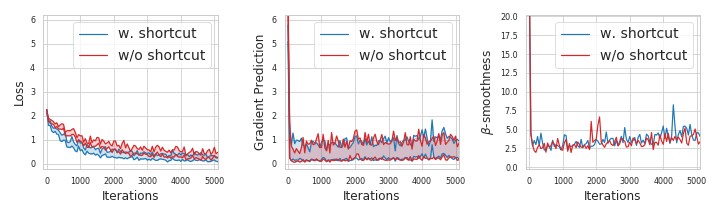}
    \caption{ResNet18 shortcut comparison.}
    \label{fig:grad_pred_shortcut_comparison}
\end{figure}

\begin{figure}[t]
    \centering
    \includegraphics[width=1.0\linewidth]{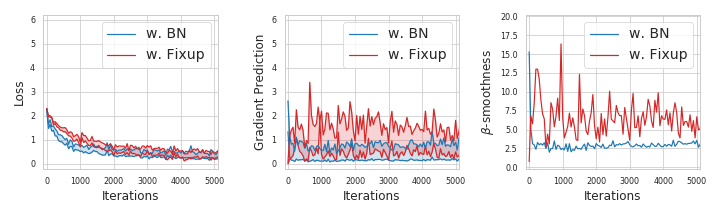}
    \includegraphics[width=1.0\linewidth]{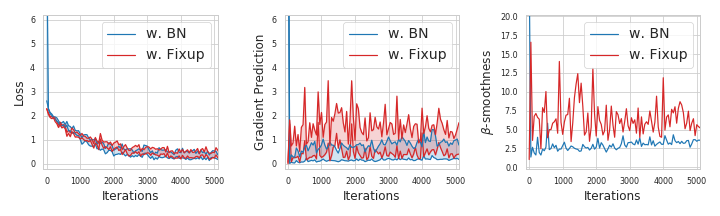}
    \caption{Comparison of BN and FixupIni on Resnet20 (top) and ResNet56 (bottom).}
    \label{fig:grad_pred_fixup}
\end{figure}

Following~\cite{santurkar2018does}, we visualize the optimization landscape. The results on ResNet18 and VGG16 are shown in Fig.\ref{fig:landscape_resnet18} and Fig.\ref{fig:landscape_vgg16}, respectively. On ResNet18, only BN leads to a more predictive and stable gradient; on ResNet50, BN/IN/LN/GN lead to a more stable gradient, however, the effect of IN/LN/GN is significantly smaller than that of BN. The results demonstrating the influence of shortcut are shown in Fig.\ref{fig:grad_pred_shortcut_comparison} where the shortcut is found to have a trivial influence on the gradient stability. The results comparing FixupIni and BN are shown in Fig.\ref{fig:grad_pred_fixup}, where FixupIni leads to a less stable gradient than BN.

\end{document}